# CG-MER: A Card Game-based Multimodal dataset for Emotion Recognition

Nessrine Farhat, Amine Bohi[*], Leila Ben Letaifa and Rim Slama
CESI LINEACT Laboratory, France

**ABSTRACT**

The field of affective computing has seen significant advancements in exploring the relationship between emotions and emerging technologies. This paper presents a novel and valuable contribution to this field with the introduction of a comprehensive French multimodal dataset designed specifically for emotion recognition. The dataset encompasses three primary modalities: facial expressions, speech, and gestures, providing a holistic perspective on emotions. Moreover, the dataset has the potential to incorporate additional modalities, such as Natural Language Processing (NLP) to expand the scope of emotion recognition research. The dataset was curated through engaging participants in card game sessions, where they were prompted to express a range of emotions while responding to diverse questions. The study included 10 sessions with 20 participants (9 females and 11 males). The dataset serves as a valuable resource for furthering research in emotion recognition and provides an avenue for exploring the intricate connections between human emotions and digital technologies.

**Keywords:** Emotion recognition, Spontaneous dataset, Multimodal data, Artificial intelligence, Affective computing.

## 1. INTRODUCTION

Emotions profoundly impact our daily experiences and decision-making processes. Recognizing and understanding emotions is crucial for improving interpersonal interactions and has diverse applications in fields such as human-computer interaction [1], mental healthcare [2], market research [3], and education [4]. These applications demonstrate the broad impact of emotion recognition in advancing various communities, including artificial intelligence, computer vision, speech processing, human sciences, and psychology.

The development and evaluation of emotion recognition models rely on the availability of diverse and high-quality unimodal datasets. In the realm of facial emotion recognition, widely used datasets include AffectNet [5] and Fer2013 [6]. For speech emotion recognition, datasets such as RAVDESS [7] and EmoDB [8] have been utilized. Moreover, we identified the ChaLearn Gesture Challenge (CGC) [9] dataset for gesture-based emotion recognition. These unimodal datasets have significantly contributed to advancing emotion recognition models within their respective modalities. However, it is important to acknowledge that unimodal datasets may not cover the complete spectrum of emotions expressed through multiple modalities [10]. To overcome this limitation, using multimodal datasets has become crucial, as they provide a comprehensive understanding of emotions. These datasets combine audio, visual, physiological signals, and other modalities, offering a holistic representation and allowing for a more in-depth exploration of emotional cues.

In this paper, we introduce CG-MER, a Card Game-based Multimodal dataset for Emotion Recognition, constructed during sessions involving an emotional card game. Our primary contribution lies in the introduction of a new benchmark accompanied by multimodal annotations. Furthermore, we present an original protocol centered on the emotion game, wherein participants engage in spontaneous and convivial conversations, expressing various emotions while responding to questions that vary in intensity.

This paper is organized into three main sections. Section 2 delves into the related work, where we explore human emotion recognition, covering both monomodal and multimodal approaches. In Section 3, we present our CG-MER dataset, providing comprehensive details on its setup, experimental procedure, data description, and annotation process. Finally, Section 4 concludes the paper, summarizing our findings and discussing the perspectives.

[*] Corresponding author : abohi@cesi.fr

## 2. RELATED WORK

In this section, we will provide a concise overview of recent studies concerning the identification of human emotions. Our primary focus will be on the utilization of multimodal datasets, which have played a vital role in evaluating the effectiveness of deep learning models in the field of emotion recognition.

### 2.1. Human emotion recognition

Emotion recognition is a prominent research area with potential impacts on understanding human behavior and enhancing human-computer interactions. It involves multiple modalities such as facial expressions, speech, gestures, and physiological signals. In the domain of facial emotion recognition, extensive surveys have been conducted, providing comprehensive overviews of techniques and advancements. Notably, Zhang et al. [11] conducted an in-depth survey that explored various facial expression recognition techniques. Speech emotion recognition has also been extensively investigated, with surveys like the one conducted by Abbaschian et al. [12] delving into feature extraction, classification techniques, and relevant databases, offering valuable insights into the domain. Furthermore, Khalil et al. [13] provided an overview of deep learning techniques and their applications in speech-based emotion recognition. Gesture-based emotion recognition has also received considerable attention, and surveys, such as the one conducted by Noroozi et al. [14], have provided valuable insights into this modality. With the advent of multimodal approaches, surveys focusing on fusion techniques and challenges have emerged as well. For instance, Koromilas et al. [15] reviewed the state-of-the-art multimodal speech emotion recognition methodologies, with a particular emphasis on integrating audio, text, and visual information, another survey [16] has presented the basic research in the field and the recent advances into the emotion recognition from facial, voice, and physiological signals, where the different modalities are treated independently. These surveys contribute significant insights into the advancements and potential of multimodal emotion recognition, highlighting the benefits of leveraging multiple modalities to enhance accuracy and robustness in emotion recognition models.

### 2.2. Emotion recognition datasets

This subsection provides an overview of well-known multimodal datasets that have been widely utilized for multimodal emotion recognition in the literature.

In [17], the authors introduced the K-EmoCon dataset, which encompasses videos, vocal audio, and biosignals recorded during debates involving 32 individuals. This dataset covers 5 emotional classes. Another notable dataset, AMIGOS [18], focuses on group affect, personality, and mood. It comprises facial expressions, audio recordings, and physiological responses collected from 40 participants across 7 emotional classes. LUMED [19], a multimodal dataset capturing simultaneous audiovisual data from 13 participants. The dataset includes selected web-based video clips and covers 3 basic emotional classes. The MELD dataset was presented in [20], featuring dialogue excerpts from the sitcom Friends along with textual, audio, and video recordings. This dataset involves 7 participants and spans 7 emotional classes. IEMOCAP [21] introduced by Busso et al., offers audiovisual and motion data from dyadic sessions involving 10 actors with 5 emotional classes. Additionally, the EMOFBVP dataset, outlined in [22], captures the expressions of 10 actors through facial and vocal expressions, body gestures, and physiological signals, covering 23 different emotional classes. In addition, the MSP-IMPROV dataset, presented by Caridakis et al. [23], that serves to study non-verbal behavior across different cultures specifically, German, Greek, and Italian. The dataset comprises 51 participants who express emotions through gestures and speech while reading Velten sentences. Recently, a large audio-visual dataset denoted Empathic [24] has been developed. This dataset involves 250 speakers from three European countries namely Spain, France and Norway.

Although the reviewed datasets have contributed significantly to the field of emotion recognition, they also exhibit certain limitations that warrant further improvements. Some datasets may lack enough participants to train deep learning models effectively, while others might not contain all the necessary modalities with annotated data for robust model training. Additionally, limited classes in some datasets might not cover the full spectrum of emotional states. Furthermore, the absence of a suitable French dataset with adequate simplicity and favorable conditions for usage is also a concern. In Table 1, we present a comprehensive comparison between our dataset and existing ones, considering factors like participant count, modalities, experiments, annotation methods, and number of classes. Understanding these strengths and limitations allows us to recognize the unique contributions and potential of our dataset in advancing multimodal emotion recognition research.

Table 1. Comparison of existing multimodal datasets for emotion recognition with our CG-MER dataset: A comprehensive analysis of participants, modalities, experiments, annotation methods, and number of classes, while N.M = Not Mentioned, Spon. = spontaneous, Both = spontaneous and posed. For the context, I = Individual, D = Dyadic, and G = Group. For annotation types, S = Self-annotations, P = Partner annotations, and E = External observer annotations.

| Dataset [ref] | Context Type | Participants Number | Gender | Age | Language | Actor/No | Modalities Visual Face | Gesture | Voice | Signals | NLP | Data Experiment | Spon/posed | # Emotions | Annotation Method | Type |
|---|---|---|---|---|---|---|---|---|---|---|---|---|---|---|---|---|
| K-EmoCom (2015) | D | 32 | 12f, 20m | [19,36] | English | No | √ | | √ | √ | | Debate | Spon | 5 | S,P,E | Interval-based continuous Seq of 5 sec |
| AMIGOS (2017) | I,G | 40 | 13f, 27m | [21,40] | English | No | √ | √ | √ | √ | | Watching videos | Spon | 7 | S,E | Per stimuli |
| LUMED-2 (2018) | I | 13 | 7f, 6m | N.M | English | No | √ | | | √ | | Watching videos | Spon | 3 | S | Per stimuli |
| IEMOCAP (2019) | D | 10 | 5f, 5m | N.M | English | Yes | √ | √ | | | | Dyadic sessions | Both | 5 | S,E | Per dialog turn |
| MELD (2019) | D,G | 7 | 3f, 3m +o | Age of actors | English | Yes | √ | | √ | | √ | Collect videos | Both | 7 | E | Turn-based |
| EMOFBVP (2020) | I | 10 | N.M | N.M | English | yes | √ | √ | √ | | | Ask actors to display emotions | Spon | 23 | E | Per stimuli |
| MSP-IMPROV (2020) | I | 11 | 5f, 6m | [23,40] | Greek | No | √ | √ | √ | | | Velten phrases | Spon | 3 | E | Per stimuli |
| | | 21 | 10f, 11m | [20,28] | German | | | | | | | | | | | |
| | | 19 | 8f, 11m | [24,48] | Italian | | | | | | | | | | | |
| Empathic (2021) | I | 134 | N.M | [64,79] | Spanish | No | √ | √ | | | | Coaching | Spon | 6 | S,E | Turn-based |
| | | 76 | N.M | [64,79] | French | | | | | | | | | | | |
| | | 62 | N.M | [64,79] | Norwegian | | | | | | | | | | | |
| **CG-MER** | **D** | **20** | **9f, 11m** | **[20,43]** | **French** | **No** | **√** | **√** | **√** | | | ***Emotional card game*** | ***Spon*** | **7** | ***S, P, E*** | ***Per question/ Per time range*** |

## 3. THE PROPOSED DATASET: CG-MER

The challenge with existing emotion datasets lies in their limited applicability to real-world contexts, due to induced emotions in controlled environments. Unlike typical datasets that use specific stimuli, CG-MER dataset employs an emotional card game to explore genuine emotions between participants. Its goals are to investigate correlations between emotional stimuli and expressions, understand emotions in social contexts, and capture dynamic emotional experiences. The dataset adheres to ethical standards approved by GDPR (General Data Protection Regulation) in France, with participants providing written consent after reviewing detailed information about data collection, privacy, and data deletion options.

### 3.1. Data collection setup

The dataset comprises 10 sessions involving 20 participants, including 9 females and 11 males. The age range of the participants spans from 20 to 43 years old, ensuring a diverse representation within the dataset. To introduce the study and emphasize the importance of constructing a multimodal dataset for emotion recognition, we prepared an engaging animation campaign. This campaign served as an introduction, presenting the research topic and emphasizing the significance of developing a multimodal dataset for emotion recognition. To facilitate participant involvement, a structured protocol was implemented. Participants were asked to complete a form, allowing them to choose their preferred partner and select their desired participation time slot. A pre-arranged schedule facilitated the organization and coordination of the participant's involvement in the study. All data collection sessions took place in a dedicated room with the same temperature and illumination conditions. To ensure optimal data capture, two participants were seated across a table, maintaining a comfortable distance between them to facilitate communication (Figure 1). To capture facial expressions and movements, two Kinect cameras V1 were positioned at the center of the table, facing each participant.

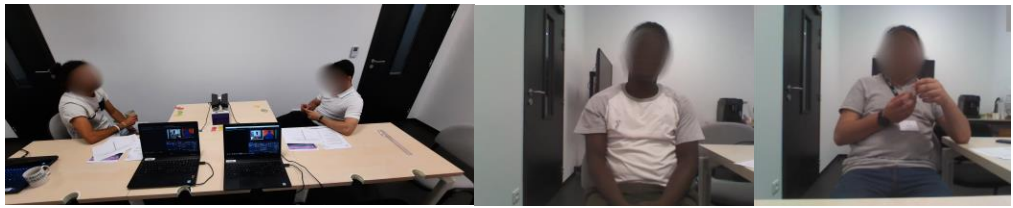

Figure 1. The left image depicts a pair of participants engaged in the card game while seated at a table. Two Kinect cameras positioned in the middle of the table were used to capture the participants' facial expressions, upper body movements and voice, as illustrated on the right side in the sample screenshot of the recorded footage.

To configure the cameras, we installed KinectSDK-v1.8, which included the Kinect driver and runtime for the Windows Software Development Kit (SDK). This SDK enabled developers to create applications that support face, gesture, and voice recognition utilizing the Kinect sensor technology. Additionally, we utilized the Kinect for Windows Developer Toolkit, which provided valuable resources such as source code samples and tools, streamlining the development of Kinect for Windows applications. During the data collection sessions, we utilized OBS Studio (Version: 29.1.1)[1] to capture both color and depth videos for each participant, allowing us to capture facial expressions and gestures accurately. For audio capture, we employed the Audio Capture Row-Console application available in the Kinect developer kit for each participant.

### 3.2. Data collection procedure

The data collection procedure consisted of three steps to ensure a systematic and organized process. Two experimenters administered each session to configure the cameras and software, to explain the game to participants, and to manage the record during the session.

**Step 1:** Consent and data usage agreement

Upon arrival, each participant was provided with a consent form outlining the purpose of the study and the use of their data, including facial expressions, voice, and gestures, for research purposes. They were given ample time to review the form and provide their written consent for participating in the data collection process.

**Step 2:** Game explanation and annotation instructions

Following the consent process, the participants were briefed about the card game and its rules. They were given a detailed explanation of how the game would unfold and were encouraged to ask any questions for clarification. Additionally, they were introduced to the annotation form, which contained sections for self-annotation and partner annotation for the seven predefined emotions (happiness, anger, neutral, fear, sadness, surprise, and disgust). The participants were guided on how to effectively use the annotation form to record their own emotional experiences and observe their partner's emotions during the card game. Furthermore, the participants were directed to select two cards from each of the three categories (red, yellow, and green), comprising 30 different questions, with 10 questions in each card category (examples of the questions are provided in Table 2). Each category was associated with a specific color and score. The green cards, indicated by a score of +1, represented questions with lower intensity and less personal nature. These cards contained simpler and non-personal questions. The yellow cards, marked with a score of +3, had a moderate intensity level and included questions that were slightly more personal. Lastly, the red cards, carrying a score of +5, had the highest intensity and featured questions of a more personal nature. With their selected cards in hand, encompassing a variety of colors and corresponding question types, the participants were prepared to commence the game.

---

[1] https://obsproject.com/fr

**Step 3:** Gameplay and emotion annotation

During this step, the participants actively engaged in the card game, creating a convivial and spontaneous atmosphere. Unlike a mechanized question-and-answer process, participants had the opportunity to ask additional questions to gain further insights into the given situation. After responding to each question, both participants were encouraged to annotate their own emotional states and provide observations regarding their partner's emotions using the provided annotation form. This interactive and dynamic process enabled the comprehensive collection of data on self-perceived emotions as well as the perceived emotions of their partner throughout the gameplay.

By following these three steps, we ensured a structured and standardized procedure for data collection during the card game sessions. This approach facilitated the systematic capture of multimodal data and enriched our dataset for further analysis and research on emotion recognition.

Table 2. Examples of questions in each card category (The text in bold represents the English translation of the original French questions).

| Card Type | Question Score | Intensity | Question |
|---|---|---|---|
| Green | 1 | Low | **What is your main center of interest, and why does it fascinate you?** *Quel est votre meilleur centre d'intérêt et pourquoi il vous passionne?* |
| Yellow | 3 | Medium | **How do you feel when you are alone and no one is talking to you?** *Comment vous sentez-vous lorsque vous êtes seul(e) et que personne ne vous parle?* |
| Red | 5 | High | **Have you ever experienced a phobia or an irrational fear?** *Avez-vous déjà été confronté(e) à une phobie ou une peur irrationnelle?* |

### 3.3. Dataset description

The CG-MER dataset comprises a comprehensive collection of multimodal data obtained from 10 distinct sessions involving an emotional card game, resulting in a total duration of approximately 10 hours of dyadic interactions. Each session was recorded in the form of two audiovisual recordings, saved as mp4 files. These recordings encompassed the RGB video and depth video of each participant, enabling a detailed analysis of facial expressions and body movements. To ensure focused analysis, the videos were carefully cropped using Adobe Premium software to extract the RGB and depth videos as separate entities. Furthermore, continuous annotations of emotions were collected from three distinct perspectives of the subject, the partner, and the external observers. This multimodal approach allowed for a comprehensive understanding of the emotional dynamics at play during card game interactions. In addition to the visual data, audio tracks capturing the participants' speeches were recorded, then saved as individual WAV files. This audio component provides valuable insights into the verbal expressions and vocal cues associated with different emotional states. To provide an overview of the dataset's contents and data collection outcomes, Table 3 summarizes the key details, including the number of sessions, duration, and the various data modalities captured.

Table 3. A summary of the data collection.

| Data | collection summary |
|---|---|
| Number of participants | 20 (11 males, 9 females) |
| Participants age | 20 to 43 (mean = 31,5) |
| Game sessions duration | ~30 min |
| Emotion annotation categories | Surprise, Happiness, Neutral, Anger, Sadness, Fear, Disgust |
| Game sessions videos | 10h 1min 24sec (of 20 participants) |
| Game sessions audios | 5h 3min 45sec (of 10 sessions) |
| Emotions annotations | Self: per question (120 questions for 10 sessions) Partner: per question (120 questions for 10 sessions) 3 external observers: per time range |

For the extraction of skeleton data from video, we employed the Blazepose algorithm [25], which demonstrated superior performance compared to videoPose3D as demonstrated in [26]. Blazepose allowed us to accurately capture and analyze the participants' skeletal movements by detecting and tracking multiple joints. With the ability to capture data from various joints, such as the head, neck, shoulders, elbows, wrists, hips, knees, and ankles, we ensured a comprehensive representation of the participants' movements during the emotional card game sessions.

### 3.4. Data annotation

Prior studies have demonstrated that incorporating multiple sources for emotion annotations, as evidenced by [27, 28], can significantly enhance the accuracy of emotion recognition systems. The use of multiple perspectives, including those of the subject, interacting partner, and external observers, is essential for establishing a comprehensive ground truth in emotion annotations. With the CG-MER dataset, we aim to address these complexities by embracing all three available perspectives in the annotation of emotions within the context of social interactions.

**Self and Partner Annotation:** During the interactive gameplay sessions of the emotional card game, participants were involved in self and partner annotation. After responding to each question, both participants were given approximately 1 minute to annotate their own emotional experiences and those of their partner. They utilized a predefined table containing the 7 basic emotions, allowing them to express their feelings and observations in real-time. The self-annotation provided insights into the subject's immediate emotional responses, while the partner annotation captured the perceived emotions experienced by the interacting partner during the dyadic interactions.

**External Observers Annotation**: Three external raters were recruited for this purpose. Using "label-studio", the open-source data labeling tool. For the external observers' annotation, we adopted a data-driven approach that relies on both the self and partner annotations provided during the game sessions. These annotations served as the basis for external observers to assess and annotate the participants' emotions.

| SESSION | Session 1 | | Session 2 | | Session 3 | | total |
|---|---|---|---|---|---|---|---|
| Participant | P1 | P2 | P1 | P2 | P1 | P2 | - |
| Happiness | 5 | 3 | 3 | 1 | 2 | 4 | 18 |
| Surprise | 4 | 8 | 4 | 2 | 4 | 1 | 23 |
| Neutral | 13 | 21 | 12 | 12 | 10 | 6 | 74 |
| Anger | - | 7 | 2 | 6 | 1 | - | 16 |
| Sadness | 1 | 1 | 2 | 5 | 1 | - | 10 |
| Fear | 3 | 1 | - | - | - | - | 4 |
| Disgust | - | 3 | 3 | - | - | - | 6 |
| total | 26 | 44 | 26 | 26 | 18 | 11 | 151 |

Table 4. Example of an external observer's annotation for the first three sessions

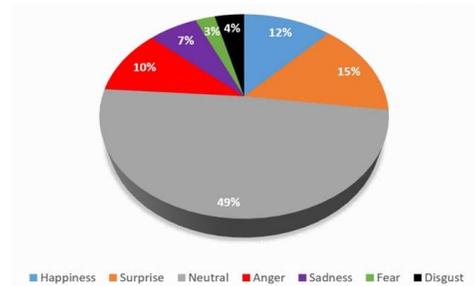

Figure 2. Emotion distribution across participants in some selected sessions

By leveraging the combination of self and partner annotations along with the impartial observations of external raters, we aimed to capture a more holistic and robust representation of the participants' emotional experiences. Table 4 presents examples of an external observer's annotation for some sessions, while Figure 2 illustrates the emotion distribution across participants in selected sessions.

### 3.5. Final dataset

The final dataset encompasses a rich collection of multimodal data from 10 distinct sessions of emotional card games. These sessions capture genuine and spontaneous interactions between participants, offering valuable insights into emotional expressions and perception during social interactions. The CG-MER dataset is organized into ten folders, each corresponding to a specific session of the emotional card game. Within each session folder, there are ten individual files, each representing a participant's data. For each participant, the dataset includes the following data files:

**RGB Video (mp4):** A video file capturing the participant's facial expressions and gestures during the game session.

**Depth Video (mp4):** A video file providing in-depth information of the participant's body movements.

**Audio (wav):** An audio file capturing the participant's speech and vocal expressions.

**Skeleton Video (mp4):** A video file visualizing the skeletal movements of the participant during the game session.

**External Observers' Annotations (json, csv):** Two separate files in JSON and CSV formats containing the annotations provided by external observers who rated the participants' emotional states during the interactions.

The dataset's multimodal nature enables researchers to delve into various aspects of emotion recognition, incorporating facial expressions, vocal cues, and gestures to gain a comprehensive understanding of emotional dynamics during social interactions.

## 4. CONCLUSION

In conclusion, the CG-MER dataset presents a valuable resource for emotion recognition research among French-speaking individuals. While not publicly available at this time, its potential impact on affective computing is promising, fostering interdisciplinary research and advancing our understanding of human emotions and social interactions in diverse cultural contexts. In future work, a primary focus of our research will be to investigate the acquisition of additional data from individuals with neurodegenerative diseases to augment the second version of the CG-MER dataset. By collecting data in this specific context, we aim to gain deeper insights into the emotional responses and social interactions of individuals affected by neurodegenerative disorders.

**Acknowledgment:** We would like to express our gratitude to all the participants who contributed to the completion of this study and the development of the CG-MER database. We also extend our thanks to the external observers who assisted in annotating our dataset.

## REFERENCES

[1] M. KalpanaN. T. N. H. D. J. Chowdary, "Deep learning-based facial emotion recognition for human–computer inter- action applications," Neural Computing and Applications (2021).

[2] E. E. Lee, J. Torous, M. De Choudhury, C. A. Depp, S. A. Graham, H.-C. Kim, M. P. Paulus, J. H. Krystal, and D. V. Jeste, "Artificial intelligence for mental health care: Clinical applications, barriers, facilitators, and artificial wisdom," Biological Psychiatry: Cognitive Neuroscience and Neuroimaging 6(9), 856–864 (2021).

[3] N. Wirth, "Hello marketing, what can artificial intelligence help you with?," International Journal of Market Research 60(5), 435–438 (2018).

[4] L. Chen, P. Chen, and Z. Lin, "Artificial intelligence in education: A review," IEEE Access 8, 75264–75278 (2020).

[5] A. Mollahosseini, B. Hasani, and M. H. Mahoor, "Affectnet: A database for facial expression, valence, and arousal computing in the wild," IEEE Transactions on Affective Computing 10(1), 18–31 (2017).

[6] I. J. Goodfellow, D. Erhan, P. L. Carrier, A. Courville, M. Mirza, B. Hamner, W. Cukierski, Y. Tang, D. Thaler, D.-H. Lee, and others, "Challenges in representation learning: A report on three machine learning contests," in Neural Information Processing: 20th International Conference, ICONIP 2013, Daegu, Korea, November 3-7, 2013. Proceedings, Part III 20 , 117–124, Springer (2013).

[7] S. R. Livingstone and F. A. Russo, "The ryerson audio-visual database of emotional speech and song (ravdess): A dynamic, multimodal set of facial and vocal expressions in north american english," PLOS ONE 13, 1–35 (05 2018).

[8] F. Burkhardt, A. Paeschke, M. Rolfes, W. F. Sendlmeier, B. Weiss, and others, "A database of german emotional speech.," in Interspeech, 5, 1517–1520 (2005).

[9] S. Escalera, J. Gonza`lez, X. Baro´, M. Reyes, O. Lopes, I. Guyon, V. Athitsos, and H. Escalante, "Multi-modal gesture recognition challenge 2013: Dataset and results," in Proceedings of the 15th ACM on International conference on multimodal interaction, 445–452 (2013).

[10] M. F. H. Siddiqui, P. Dhakal, X. Yang, and A. Y. Javaid, "A survey on databases for multimodal emotion recognition and an introduction to the viri (visible and infrared image) database," Multimodal Technologies and Interaction 6(6), 47 (2022).

[11] T. Zhang, "Facial expression recognition based on deep learning: a survey," in Advances in Intelligent Systems and Interactive Applications: Proceedings of the 2nd International Conference on Intelligent and Interactive Systems and Applications (IISA2017), 345–352, Springer (2018).

[12] B. J. Abbaschian, D. Sierra-Sosa, and A. Elmaghraby, "Deep learning techniques for speech emotion recognition, from databases to models," Sensors 21(4), 1249 (2021).

[13] R. A. Khalil, E. Jones, M. I. Babar, T. Jan, M. H. Zafar, and T. Alhussain, "Speech emotion recognition using deep learning techniques: A review," IEEE Access 7, 117327–117345 (2019).

[14] F. Noroozi, C. A. Corneanu, D. Kamińska, T. Sapiński, S. Escalera, and G. Anbarjafari, "Survey on emotional body gesture recognition," IEEE transactions on affective computing 12(2), 505–523 (2018).

[15] P. Koromilas and T. Giannakopoulos, "Deep multimodal emotion recognition on human speech: A review," Applied Sciences 11(17), 7962 (2021).

[16] N. Sebe, I. Cohen, T. Gevers, and T. S. Huang, "Multimodal approaches for emotion recognition: a survey," in Internet Imaging VI, 5670, 56–67, SPIE (2005).

[17] C. Y. Park, N. Cha, S. Kang, A. Kim, A. H. Khandoker, L. Hadjileontiadis, A. Oh, Y. Jeong, and U. Lee, "K-emocon, a multimodal sensor dataset for continuous emotion recognition in naturalistic conversations," Scientific Data 7(1), 293 (2020).

[18] J. A. Miranda-Correa, M. K. Abadi, N. Sebe, and I. Patras, "Amigos: A dataset for affect, personality and mood research on individuals and groups," IEEE Transactions on Affective Computing 12(2), 479–493 (2018).

[19] Y. Cimtay, E. Ekmekcioglu, and S. Caglar-Ozhan, "Cross-subject multimodal emotion recognition based on hybrid fusion," IEEE Access 8, 168865–168878 (2020).

[20] S. Poria, D. Hazarika, N. Majumder, G. Naik, E. Cambria, and R. Mihalcea, "Meld: A multimodal multi-party dataset for emotion recognition in conversations," arXiv preprint arXiv:1810.02508 (2018).

[21] C. Busso, M. Bulut, C.-C. Lee, A. Kazemzadeh, E. Mower, S. Kim, J. N. Chang, S. Lee, and S. S. Narayanan, "Iemocap: Interactive emotional dyadic motion capture database," Language resources and evaluation 42, 335–359 (2008).

[22] H. Ranganathan, S. Chakraborty, and S. Panchanathan, "Multimodal emotion recognition using deep learning architectures," in 2016 IEEE winter conference on applications of computer vision (WACV), 1–9, IEEE (2016).

[23] G. Caridakis, J. Wagner, A. Raouzaiou, F. Lingenfelser, K. Karpouzis, and E. Andre, "A cross-cultural, multimodal, affective corpus for gesture expressivity analysis," Journal on Multimodal User Interfaces 7, 121–134 (2013).

[24] L. B. Letaifa and M. I. Torres, "Perceptual borderline for balancing multi-class spontaneous emotional data," IEEE Access 9, 55939–55954 (2021).

[25] V. Bazarevsky, I. Grishchenko, K. Raveendran, T. Zhu, F. Zhang, and M. Grundmann, "Blazepose: On-device real-time body pose tracking," arXiv preprint arXiv:2006.10204 (2020).

[26] S. Deb, M. F. Islam, S. Rahman, and S. Rahman, "Graph convolutional networks for assessment of physical rehabilitation exercises," IEEE Transactions on Neural Systems and Rehabilitation Engineering 30, 410–419 (2022).

[27] J. Healey, "Recording affect in the field: Towards methods and metrics for improving ground truth labels," in Affective Computing and Intelligent Interaction: 4th International Conference, ACII 2011, Memphis, TN, USA, October 9–12, 2011, Proceedings, Part I 4, 107–116, Springer (2011).

[28] B. Zhang, G. Essl, and E. Mower Provost, "Automatic recognition of self-reported and perceived emotion: Does joint modeling help?," in Proceedings of the 18th ACM International Conference on Multimodal Interaction, 217–224 (2016).